\documentclass[9pt]{article}
\usepackage{spconf,amsmath,graphicx}

\usepackage[utf8]{inputenc} % allow utf-8 input
\usepackage[T1]{fontenc}    % use 8-bit T1 fonts
\usepackage{hyperref}       % hyperlinks
\usepackage{url}            % simple URL typesetting
\usepackage{booktabs}       % professional-quality tables
\usepackage{amsfonts}       % blackboard math symbols
\usepackage{nicefrac}       % compact symbols for 1/2, etc.
\usepackage{microtype}      % microtypography
\usepackage{xcolor}         % colors

\usepackage{amssymb}
\usepackage{mathtools}
\usepackage{bbm}
\usepackage{cleveref}
\usepackage{amsthm}
\usepackage{pifont}
\newcommand{\cmark}{{\color{green} \ding{51}}}
\newcommand{\xmark}{{\color{red} \ding{55}}}
\usepackage{algorithm}
\usepackage[noend]{algpseudocode}
\usepackage{multicol}
\usepackage{wrapfig}

\newcommand{\R}{\mathbb{R}}
\newtheorem{theorem}{Theorem}
\newtheorem{lemma}{Lemma}

\title{OTW: Optimal Transport Warping for Time Series}
\name{Fabian Latorre$^{\star}$\thanks{All work done while the first author was an intern at Salesforce Research.}, Chenghao Liu$^{\dagger}$, Doyen Sahoo$^{\dagger}$, Steven C.H. Hoi$^{\dagger}$}
\address{$^{\star}$École polytechnique fédérale de Lausanne (EPFL) \\ $^{\dagger}$Salesforce Research Asia}
\begin{document}
%\ninept
%
\maketitle
\begin{abstract}
Dynamic Time Warping (DTW) has become the pragmatic choice for measuring distance between time series. However, it suffers from unavoidable quadratic time complexity when the optimal alignment matrix needs to be computed exactly. This hinders its use in deep learning architectures, where layers involving DTW computations cause severe bottlenecks. To alleviate these issues, we introduce a new metric for time series data based on the Optimal Transport (OT) framework, called Optimal Transport Warping (OTW). OTW enjoys linear time/space complexity, is differentiable and can be parallelized. OTW enjoys a moderate sensitivity to time and shape distortions, making it ideal for time series. We show the efficacy and efficiency of OTW on 1-Nearest Neighbor Classification and Hierarchical Clustering, as well as in the case of using OTW instead of DTW in Deep Learning architectures. 
\end{abstract}
\section{Introduction}
\label{sec:introduction}
 Time-series data is ubiquitous in contemporary Machine Learning applications including Heart disease predictions based on ECG data \cite{TRIPOLITI201726}, end-to-end text-to-speech synthesis \cite{Donahue2021} and sign-language recognition
\cite{signlanguage}. Common tasks like supervised classification and hierarchical clustering require a notion of \textit{distance} or \textit{similarity} between time series, and their performance strongly depends on such a choice. A key desired characteristic of a time series distance
is the ability to identify commonly occurring patterns such as shape distortions and time delays. For example, two time series that differ by a slight distortion in shape or a slight delay in time should be considered \textit{similar} or \textit{close} \cite{accelerometer}.

The euclidean distance is unable to reason about shape and time distortions, and thus, is considered a poor choice for time series applications. Dynamic Time Warping (DTW) \cite{Sakoe1978}can assign high similarity values to time series that have a closely matching shape, but which do not necessarily align perfectly in the time domain. For this reason, DTW has established itself as the most
prominent similarity measure for time series  \cite{dtw1,dtw2,Keogh2004,Cuturi2017,Cai2019}.

However, DTW and all closely-related variants suffer from an unavoidable quadratic
complexity  with respect to the length of the time series \cite{Abboud2015}. This makes it highly expensive or outright unusable for time series of considerable length, especially for large datasets. This issue has been noted in literature, and researchers have developed faster variants of DTW, such as FastDTW\cite{Salvador2004}. Unfortunately, FastDTW has been dismissed as being slower than DTW in practice, despite its theoretical linear time complexity \cite{Wu2020FastDTWIA}.

Recently, DTW has been incorporated inside Deep Learning pipelines for time-series data. It has been used either as a loss function \cite{Leguen2019,Chang2019}, or as a replacement for
linear layers as in DTWNet \cite{Cai2019}. In the latter case it replaces the inner product operation in the linear layers of the network. Typically, these layers have a linear time-complexity with regard to input size (e.g. convolutional layers, batch-norm, etc.), and replacing them with a quadratic complexity DTW layer introduces a significant computational bottleneck.

Moreover, DTW uses Dynamic Programming, an inherently sequential framework that does not lend itself to trivial paralellization in GPU.  Even though there are available GPU implementations \cite{Schmidt2020,Maghoumi2020,Lee2021}, they do not avoid the sequential nature. This limits the adoption of DTW in practice, as the cost of training and hyperparameter optimization increases considerably. Lastly, DTW is not a metric, as it violates the triangle inequality. Thus, it cannot exploit faster similarity search methods like the Approximating and Eliminating Search Algorithm (AESA) \cite{Vidal1986}. 

Thus, we need a distance notion that not only can be computed in linear time, but is also differentiable and can be parallelized on GPU, thus speeding up all training pipelines by a huge margin. Even though the euclidean distance enjoys such characteristics, it does not provide a good inductive bias for time-series data, as it ignores the time component. Precisely, the goal is to keep the theoretical properties and performance of DTW at a computational cost similar to that of the euclidean distance.

\textbf{Our contributions. } It is apparent that there is a need for new time series distance notions that overcomes some, if not all, the aforementioned drawbacks. We propose a
new distance for time series data that we call \textbf{Optimal Transport Warping (OTW)}. Our distance is rooted in the theory of Optimal Transport \cite{Peyre2019} (OT), which is a well-known metric used to compare the shape of two probability distributions.

We adapt Optimal Transport for the case of time-series data through an Unbalanced Optimal Transport formulation, given that two time series may not have the same mass (may not sum up to the same value). We also address the issue of negativity, i.e., while probability distributions are non-negative, time series may not be.
%%%We show that OTW can be computed in linear time, and also that the OTW distance increases linearly with time-shift.
Our final OTW formulation (1) can be computed in linear time and space, (2) is differentiable, (3) can be easily computed on massively parallel architectures (GPU/TPUs), (4) has moderate sensitivity to shape and time distortions, and (5) is a proper metric. A comparison of our method against several state-of-the-art time series distances is summarized in Table \ref{tab:comparison-distances}.

In experiments we observe that OTW runs up to 10-30x times faster than DTW, depending on the dataset. In the classification task it improves over DTW in 6 out of 7 types of datasets in the UCR time-series benchmark (\cref{tab:1nn-rank-comparison}). In the clustering task, it improves over DTW in all 7 types of datasets considered (\cref{tab:clustering-comparison}). A total of 92 datasets were considered. Through synthethic and real experiments we show that by replacing DTW with OTW in DTW-Net we solve its computational bottleneck, and we can achieve a lower error in less than 50\% of the time.

\section{Related Work}
\label{sec:related}
We summarize the most prominent time-series distances in
\cref{tab:comparison-distances}, and conclude that there are few low-complexity alternatives to DTW, which suffers from quadratic complexity. Only GDTW\cite{Xiang2021}, FastDTW \cite{Salvador2004} and OTW (our proposed method) are fast alternatives that avoid computational bottlenecks. However, FastDTW has been dismissed by the community \cite{Wu2020FastDTWIA} while GDTW only provides an CPU implementation of a stochastic rather than a deterministic gradient. Because the activations appear at every layer, this prevents the computation of an unbiased estimator of the gradient of the loss function. Moreover, GDTW requires a Contiguous Sparse Matrix Data Structure that might not be efficient in GPU (the authors of the paper do not provide a GPU implementation). Hence, a priori we expect that training networks using GDTW activations on CPU is an unfeasible task given limited time. 
\begin{table}[h]
\begin{center}
\begin{tabular}{ l c c c  }
\textbf{Method}                       & \textbf{Complexity}    & \textbf{Gradient} & \textbf{GPU}    \\  \hline
TWED \cite{Marteau2009}         & {\color{red} $O(n^2)$} & \xmark            & \xmark          \\
ERP \cite{Chen2004}            & {\color{red} $O(n^2)$} & \xmark            & \xmark           \\
EDR \cite{Chen2005}            & {\color{red} $O(n^2)$} & \xmark            & \xmark         \\
DTW \cite{Sakoe1978}           & {\color{red} $O(n^2)$} & \xmark            & \cmark           \\  
SoftDTW \cite{Cuturi2017}       & {\color{red} $O(n^2)$} & \cmark            & \cmark           \\
DILATE \cite{Leguen2019}          & {\color{red} $O(n^2)$} & \cmark            & \cmark           \\
SoftDTW div. \cite{Blondel2020}   & {\color{red} $O(n^2)$} & \cmark            & \cmark         \\
FastDTW \cite{Salvador2004}    & {\color{green} $O(n)$} & \xmark            & \xmark          \\
GDTW \cite{Xiang2021}             & {\color{green} $O(n)$} & Stochastic            & \xmark          \\
% $\ell^p_p$-norm                       & {\color{green} $O(n)$} & \cmark            & \cmark         & \cmark  \\
OTW (This work)                        & {\color{green} $O(n)$} & \cmark            & \cmark           \\ \hline
% \textbf{Barycenter}
% \xmark             
% \xmark             
% \xmark             
% \xmark             
% \xmark             
% \xmark             
% \xmark             
% \xmark             
% \xmark             
% \cmark             
% \cmark             
\end{tabular} \vspace{0.1cm}
\caption{Comparison of time series similarity measures. \textbf{Gradient:}
	differentiability with respect to its inputs.
	\textbf{GPU:} existence of a parallel GPU implementation.}
        \label{tab:comparison-distances}
	%\textbf{Average:} whether there exists an efficient method to
	%compute the barycenter of a set of points with respect to the
	%similarity measure. 
\end{center}
\end{table}

In recent years, DTW has found popularity in several applications using Deep Learning architectures. DTW-like
distances have been used either as a loss function for time series
forecasting/regression, or as part of a feature-extracting module that can be
used as a layer inside a neural network. When such distances are used to
replace inner products, as in linear layers or transformers, the complexity is
inevitably increased. Because layers are applied in a sequential fashion (as they cannot be parallelized), this forms a bottleneck that slows down the learning pipeline.

A few recent works developing Deep Learning pipelines that make use of a DTW-like module, potentially suffering from speed issues: 
DILATE \cite{Leguen2019} is a loss function module computing Soft-DTW as a subroutine; DTWNet \cite{Cai2019}      proposes a feature-extraction module based on DTW; D3TW \cite{Chang2019}     is a discriminative Loss for Action Alignment and Segmentation based on Soft-DTW; SLR \cite{Pu2019} is an Encoder-Decoder architecture with Soft-DTW alignment constraint for Continuous Sign Language Recognition; 
STRIPE \cite{LeGuen2020}    is similar to \cite{Leguen2019}; DTW-NN \cite{Iwana2020}    is similar to \cite{Cai2019}; SpringNet is a Transformer architecture with DTW replacing inner products; TTS \cite{Donahue2021} is an End-to-End text-to-speech architecture with a Soft-DTW-based prediction loss; 
\cite{Zelinka2020} uses DTW to synchronize a resultant and a target sequence inside a text-to-video sign-language synthesizer network. Due to the soft-DTW computational demands, an alternative attention mechanism is studied.

\textbf{OT-based time series distances.} The potential of Optimal Transport for time-series has been
explored in other works. For example, \cite{Janati2020} proposes to compare positive time-series using sinkhorn divergences, a variant
of OT. However, even in the one-dimensional
case there exists no subquadratic algorithm for the problem. Time-adaptive Optimal Transport \cite{Zhang2020} is a similar approach. Nevertheless, their formulation is
essentially different from ours: for a time series $a$, a uniform distribution
is constructed, over pairs $(i, a_i)$, which are then compared using the
traditional OT formulation. Because this is a two-dimensional distribution, the complexity of their proposed algorithm is at least quadratic and hence, as slow as DTW. In contrast our algorithms work in linear time.  We recall that OT requires cubic time but it can be approximated in quadratic time using Sinkhorn iterations \cite{Cuturi2013}. Only in the one-dimensional case  it has a linear-time implementation \cite{Vallender1974}. Finally, \cite{Su2019} interprets the values of a time series as \textit{set} of samples rather than a sequence. This has the effect of removing the time information of the sequence and hence OT, as used in this case, is oblivious to the temporal nature of the data. The resulting optimization problems are solved with the so-called Sinkhorn iterations which again lead to a quadratic complexity. In summary, ours is the first linear-time distance for time-series based on OT.

\section{Optimal Transport Warping}
\label{sec:definition-em-dwt}

% Next, we introduce the proposed distance metric for time series: Optimal Transport Warping.

\subsection{Problem Setting and Optimal Transport}
Probability distributions are analogous to density functions in physics, which measure the concentration of mass along an infinitesimal unit of volume. We can think of certain time-series as measuring a concentration of mass along an infinitesimal length of \textit{time}: if we let $a(t)$ be the amount of car traffic at time $t$ of the day, after normalization, we can think of the integral $a(t)$ over a time interval, as the probability that a car is in traffic during that interval.

Hence, a method that compares probability distributions can potentially compare time-series. Indeed, Optimal transport has been used to compare time-series or general sequences before \cite{Cuturi2013,Su2019,Zhang2020}. Optimal transport provides a many-to-many alignment, instead of a one-to-many alignment in the case of DTW and its variants. A priori, it is reasonable to believe that some time-series data might be more amenable to alignment using one type of alignment over the other. Indeed, OT is also sometimes called the \textit{Earth Mover's Distance}, because it captures the cost of \textit{reshaping} one distribution into another. In this way it is sensitive to shape distortions, a fact that has been exploited and explained in depth in \cite{Janati2020}.

A discrete time series $a$ is a sequence of numbers
$(a_1, \ldots, a_n)$, that can be understood as a function from the
set $[n]=\{1, \ldots, n\}$ to the real numbers.  Since Optimal Transport is a distance notion between probability measures, we first focus
on time series that are positive and sum to one i.e., $\sum_{i=1}^n a_i = 1; a_i \ge 0$ and $\sum_{j=1}^n b_j = 1; b_i \ge 0$.
Later, we will expand the notion of distance to arbitrary time series. To define optimal transport for probability distributions over the set $[n]$, we require a nonnegative function  $d: [n] \times [n] \to \R_+$ defined over pairs of elements of $[n]$. We denote by
$D$ the matrix with entries $D_{i, j} \coloneqq d(i, j)$ and we refer to it as
the \textit{distance matrix}. Denoting the column vector of all-ones as $\mathbbm{1} \in \R^n$,  
the Optimal Transport distance (with respect to the distance matrix $D$) between $a$ and $b$ is
defined as:
\begin{equation}
\label{eq:ot}
W_D(a, b) \coloneqq \min_T
	\left \{ \langle T, D \rangle : T \geq 0,
  	T \mathbbm{1} = a, \mathbbm{1}^T T = b \right \}
\end{equation}
 For an introduction to Optimal Transport please refer to \cite{Peyre2019}. Note that we can extend definition \eqref{eq:ot} to sequences that sum up to the same value, not necessarily equal to $1$:
 because $a$ is defined as the row-sums of $T$, and $b$ is defined as the column-sums of $T$,  $a$ and $b$ need only sum up to the same value, equal to the sum of all the entries of $T$. However, we cannot remove the constraint that
that $a,b \ge 0$, as the entries of $T$ are positive c.f. \cref{eq:ot}.

In our case we will choose the matrix $D$ as the absolute value distance,
i.e., $d(i, j) \coloneqq |i - j|$. This choice means that transporting a unit of mass from the position $i$ to the position $j$ will be equal to the absolute difference between the two positions. In the following we will always assume that $D$ is chosen in this way and we let $W(a,b) \coloneqq W_D(a, b)$ to simplify notation. This choice of $D$ ensures that there exists a closed form solution for the Optimal Transport problem \cref{eq:ot} that can be computed in linear time
\cite{Vallender1974}:
\begin{equation}
\begin{split}
	\label{eq:closed-form}
	W(a, b) = \sum_{i=1}^n |A(i) - B(i)| \\ A(i)
		\coloneqq \sum_{j=1}^i a_j, \quad B(i) \coloneqq \sum_{j=1}^i b_j
  \end{split}
\end{equation}
that is, $A$ and $B$ are the \textit{cumulative distribution} functions of $a$ and $b$,
respectively. 

Because a time series is not necessarily a probability measure,
we cannot directly use \cref{eq:ot}. The two main issues are: (i)
\textit{unbalancedness,} as the time series may not sum up to the same value and (ii)
\textit{negativity,} as a time series may contain negative values. In order
to extend the Optimal Transport framework to arbitrary time series we need to
workaround the limitations of the definition in \cref{eq:ot}, while trying to
retain the linear space/time properties of the closed-form solution
\eqref{eq:closed-form}.

\subsection{Unbalanced Optimal Transport for time-series}
We will first assume that the sequences
$a, b$ are nonnegative but that they do not necessarily sum up to the same
value. In order to resolve the unbalancedness issue, we proceed by adding a \textit{sink}
as described in \cite[Chapter 3]{Guittet2002}:
we append one additional element to both sequences as follows:
\begin{equation}
\begin{split}
 	\label{eq:rebalance}
	\hat{a}_i \coloneqq \begin{cases*}
 		a_i & if $i \in [n]$ \\
 		\sum_{j=1}^n b_j & if $i=n+1$
 	\end{cases*} \\
 	\hat{b}_i \coloneqq \begin{cases*}
 		b_i & if $i \in [n]$ \\
 		\sum_{j=1}^n a_i & if $i=n+1$
 	\end{cases*}
  \end{split}
\end{equation}
In this way, we ensure that $\sum_i \hat{a}_i = \sum_i \hat{b}_i$. This step
can be understood as balancing the total mass of the two sequences. Now, we
also have to extend the distance matrix in some way, to account for the \textit{sink}
in the extended sequences $\hat{a}$ and $\hat{b}$. For some value $m \in
\R_+$ called the \textit{waste cost} we set:
\begin{equation}
	\label{eq:rebalance-distance}
	D(m)_{i,j} = \begin{cases*}
		|i-j| & if $i, j \in [n]$ \\
		m & if  $i=n+1, j \in [n]$ \\
		m & if  $j=n+1, i \in [n]$ \\
		0 & if  $i=n+1, j=n+1$
	\end{cases*}
\end{equation}
the idea being that any excess can be transported to the $n+1$-th point
(the sink) incurring a cost of $m$ per unit of mass. We define the \textit{(nonnegative) unbalanced Optimal Transport Distance} problem as:
\begin{equation}
\label{eq:unbalanced-ot}
\begin{split}
	\widehat{W}_{m}(a, b) &\coloneqq W_{D(m)}(\hat{a}, \hat{b}) \\ &= \min_T
	\left \{ \langle T, D(m) \rangle : T \geq 0,
	T \mathbbm{1} = \hat{a}, \mathbbm{1}^T T = \hat{b} \right \}
 \end{split}
\end{equation}
note that if the sequences $a,b$ sum up to the same value, the problem reduces
to the original \textit{balanced} optimal transport problem \cref{eq:ot}. With
this modification, it is not clear if \cref{eq:unbalanced-ot} can also be
solved in closed form in linear time and space as in \cref{eq:closed-form}.
However, we can compute an upper bound:
\begin{theorem}
	\label{thm:unbalanced-solution}
	Let $D_{i,j} = |i - j|$ and let $D(m)$ be defined as in \cref{eq:rebalance-distance}. Define
	\begin{equation}
	\label{eq:ot-distance}
	\text{OTW}_m(a, b) \coloneqq m |A(n) - B(n)| + \sum_{i=1}^{n-1} |A(i) - B(i)|
	\end{equation}
	Then, $\widehat{W}_{m}(a, b) \leq \text{OTW}_m(a, b)$. Clearly,
	$\text{OTW}_m(a, b)$ can be computed in linear time/space.
\end{theorem}
We defer the proof of \cref{thm:unbalanced-solution} to \cref{sec:proof_thm_1}.
Precisely, \textbf{we propose to use 
\cref{eq:ot-distance} as the notion of distance} between time series of positive sign. In practice, the choice
of $m<n$ works better. For large values of $n$, the choice $m=n$ would put too much weight on the first component $m |A(n) - B(n)|$, which strongly penalizes the total mass difference of the two sequences $a, b$.

As we show in \cref{lem:shift-sensitivity}, the unbalanced OT distance \eqref{eq:rebalance-distance} increases linearly when a time-shift is introduced. This makes it
ideal for time-series applications like demand forecasting, where a shift in time can represent a change in the seasonality of a product, for example. In contrast, in speech recognition where time-shifts should be ignored,
perhaps a distance like DTW might be more suitable.
\begin{lemma}
	\label{lem:shift-sensitivity}
	Let $a \in \R^n_+$ and $b$ be two time series, and let $b'$ be
	a shifted version of $b$ by $t
	< n$ units, that is $b'_i = b_{i-t}$ for $i=t+1, \ldots, n$.
	For simplicity assume that $b_{n-t+1}=\ldots=b_n=0$
	i.e., the sequence $b$ is zero-padded. In this way $\sum_i b_i = \sum_i b'_i$.
	It holds that $|\widehat{W}_m(a, b') - \widehat{W}_m(a, b)| \leq t \left (\sum_{i=1}^n a_i\right )$
\end{lemma}
The proof of \cref{lem:shift-sensitivity} is deferred to \cref{sec:proof_thm_1}.
This property means that the distance increases linearly proportional to the magnitude
of the time-shift. Note that Soft-DTW \cite{Cuturi2017} has a quadratic lower bound on its
sensitivity to time-shifts, and was presented as a main contribution in \cite[Theorem 1]{Janati2020}. In contrast we show an upper bound and our sensitivity is linear.

\textbf{Constraining the Transport map to be local. } For time series distances like
DTW, it has been observed that in practice, considering all possible
alignments might be detrimental to the performance in downstream tasks. Instead, the best choice is to constrain the alignment to perform only \textit{local} matching i.e., map samples only inside a constrained window. This is precisely the so-called DTW distance with Sakoe-Chiba band constraint c.f., \cite{Ratanamahatana2004MakingTC} for details.

Our proposed Optimal-Transport based distance is no different, as the transport map $T$ in \cref{eq:unbalanced-ot} does not have any constraint, and constitutes a many-to-many map between arbitrary positions in $[n]=\{1, \ldots, n\}$. To mitigate this issue we propose to use a windowed-cumulative-sum  rather than the full cumulative-sum $A(i)\coloneqq \sum_{j=1}^i a_j$ as originally proposed in \cref{eq:closed-form}.
More precisely we let:
\begin{equation}
\label{eq:window-cumsum}
    A_s(i)\coloneqq \sum_{j=1}^i a_j - \sum_{j=1}^{i-s} a_i, \quad B_s(i) \coloneqq \sum_{j=1}^i b_j - \sum_{j=1}^{i-s} b_j
\end{equation}
and we define the local Optimal Transport Warping distance for $ s \in \{1, \ldots, n\}$:
\begin{equation}
\label{eq:local-ot-distance}
\begin{split}
    \text{OTW}_{m, s}(a, b) \coloneqq &m | A_s(n) - B_s(n) | \\
	&+ \sum_{i=1}^{n-1} |A_s(i) - B_s(i)|
 \end{split}
\end{equation}
\textbf{Localness. } the parameter $s$ is akin to the window parameter
of the Sakoe-Chiba constraint in DTW: it interpolates between the $\ell_1$-norm distance and the Unbalanced Optimal Transport. One the one hand, the $\ell_1$-norm compares the two sequences $a$, $b$ entry-wise: it does not allow for alignment between two different time positions. On the other hand, the Unbalanced Optimal Transport distance allows transport of mass between any two time positions in the time-series. The parameter $s$ interpolates between the two extremes. In practice not all datasets require the same level of localness, and it is important to choose $s$ by cross-validation, which usually results in better performance. We summarize this fact in the following lemma, with proof deferred to \cref{sec:proof_thm_1}:

\begin{lemma}
\label{lem:interp}
For simplicity assume $m=1$. When $s=1$ then $\overline{\text{OTW}}_{1, 1}(a, b)=\|a - b\|_1$. When $s=n$ we recover the \textit{global} OTW distance \eqref{eq:ot-distance} i.e., $\text{OTW}_{1, n}(a, b) = \text{OTW}_{m}(a,b)$.
\end{lemma}

 Finally, note that $\text{OTW}_{m, s}$ is not differentiable when $A_s=B_s$, due to the presence of the absolute value function.
In order to have a fully differentiable version, it suffices to use a smooth approximation of absolute value, like the well-known \textit{smooth $\ell_1$-loss}:
\begin{equation}
L_\beta(x)=\begin{cases} x^2/(2\beta) & |x| < \beta \\
|x| - \beta/2 & |x| \geq \beta
\end{cases}
\end{equation}
We define:
\begin{equation}
    \label{eq:smooth-ot}
    \begin{split}
    \text{OTW}^\beta_{m,s}(a, b) =& m L_\beta(A_s(b) - B_s(n))\\ &+
    \sum_{i=1}^{n-1} L_\beta(A_s(i) - B_s(i))
    \end{split}
\end{equation}
and it holds that $\text{OTW}^\beta_{m, s} \to \text{OTW}_{m, s}$ as $\beta \to 0$.

\subsection{Dealing with negative values} Up to this point, we have presented a way to compare
two time series that have only positive entries, using Unbalanced Optimal Transport. However,
time series can contain negative values. In order to deal with sequences of arbitrary
sign we propose to choose one of the following strategies, using cross-validation:
\begin{enumerate}
\item Apply \cref{eq:smooth-ot} to arbitrary sequences. Note that
      this formula can be applied even in the case where some entries of the
      sequences $a$ or $b$ are negative.
\item Split arbitrary sequences $a,b$ into their positive and negative  parts i.e., $a_+=\max(a, 0)$ and $a_{-}=\max(-a, 0)$, and
      sum the unbalanced optimal transport distance between the parts of equal sign:
      \begin{equation}
      \label{eq:split-ot}
      \begin{split}
          \overline{\text{OTW}}^\beta_{m, s}(a, b) =& \text{OTW}_{m, s}^\beta(a_+, b_+) \\ &+ \text{OTW}_{m, s}^\beta(a_{-}, b_{-})
        \end{split}
      \end{equation}
\end{enumerate}

\textbf{Remark. } Because we only need to compute the windowed-cumulative-sums \eqref{eq:window-cumsum} and then apply the smooth $\ell_1$-loss to the differences $A_s(i) - B_s(i)$ for $i=1,\ldots, n$, we only need a linear number of operations, as a function of $n$. Moreover, since we only use basic operations available in Deep Learning frameworks like PyTorch, the computation can be readily performed in GPU using optimized code. Finally, because the smooth $\ell_1$-loss is convex, this implies the triangle inequality and OTW becomes a true metric. Next, we perform experiments to validate the performance of OTW.

% {\color{red}{Mention that this proposed metric achieves our goals - linear time, GPU, proper metric, etc.}}

% \section{Algorithm and Complexity}
% \label{sec:algorithm}
% \input{sections/algorithm.tex}

\section{Experiments}
\label{sec:experiments}
\subsection{Comparing OT to DTW on 1-nearest-neighbors classification}
In this experiment we train 1-nearest-neighbors classifiers on the UCR
time series classification archive, which consists of a large number of
univariate time series data. Please note that \textbf{the purpose is not to show state-of-the-art
performance}. Rather, our goal is to perform an apples-to-apples comparison of the Dynamic Time Warping (DTW) distance and our proposed OTW distance.

To choose hyperparameters for the OTW distance, for each combination we follow a 80/20 random split of the training/validation sets,
and choose the one that maximizes accuracy on the validation set. For the best hyperparameters found, we evaluate the method on the testing set. The accuracy for the 1-nearest-neighbors classification for the DTW distance with learned warping window is directly obtained from the UCR time series classification benchmark website\footnote{\url{https://www.cs.ucr.edu/~eamonn/time_series_data_2018/}}.

For our method, we do 10 independent runs and obtain a 95\% confidence interval for the test error. Because of the lower complexity of our method vs DTW (linear vs quadratic complexity), it runs considerably faster and hence we consider it a better method if it can attain the same performance as DTW. Because there is only a single testing observation for the DTW methods available in the reported benchmarks, we consider that our method is better if the confidence interval for the test error of OTW contains or is below the reported test error for DTW.

In \cref{tab:1nn-rank-comparison} we summarize the results according to the type of dataset. We group the ECG, EOG, HRM and Hemodynamics types of datasets under the name \textit{Medical}, as they contain a relatively small number of datasets. We observe a noticeable improvement in all the considered categories except the \textit{sensor} category.

Note that we remove the datasets corresponding to the motion and trajectory type, as the OTW distance is not suited for such data. OT is suited to time series that can be interpreted as probability distributions or, more generally, measures. Time-series that track the position of an object through time, do not fall in this category.  The results for each single dataset are collected in \cref{sec:nn-full-comparison}.
Overall, we observe in \cref{tab:1nn-rank-comparison} that \textbf{OTW improves over DTW on 6 out of 7 types of datasets}. On some datasets like Medical and Traffic, the advantage is apparent. In total, we evaluated on 92 dataset from the UCR time series benchmark, only the ones with missing values or variable length of sequence were discarded.

\begin{table}
\centering
\scriptsize
\caption{
            Average test error of 1-NN classification using Learned
            DTW or Learned OTW (OTW error) and number of datasets in the
            collection where OTW outperforms DTW
            }
\label{tab:1nn-rank-comparison}
\begin{tabular}{cccccc}
\toprule
\textbf{Type} & \textbf{DTW error} &         \textbf{OTW error} &  \textbf{Improv.} &  \textbf{Total} & \textbf{\%} \\
\midrule
        Image &               0.25 &  $\mathbf{0.25 \pm 0.02}$ &                    19 &              31 &    $\mathbf{61.29}$ \\
      Spectro &               0.26 &  $\mathbf{0.24 \pm 0.03}$ &                     6 &               8 &    $\mathbf{75.00}$ \\
       Sensor &               0.23 &           $0.44 \pm 0.03$ &                     9 &              28 &             $32.14$ \\
       Device &               0.34 &           $0.44 \pm 0.03$ &                     5 &               9 &    $\mathbf{55.56}$ \\
 Medical &               0.38 &  $\mathbf{0.36 \pm 0.02}$ &                     9 &              13 &    $\mathbf{69.23}$ \\
      Traffic &               0.12 &  $\mathbf{0.06 \pm 0.02}$ &                     2 &               2 &   $\mathbf{100.00}$ \\
        Power &               0.08 &  $\mathbf{0.04 \pm 0.01}$ &                     1 &               1 &   $\mathbf{100.00}$ \\
\bottomrule
\end{tabular}
\end{table}

\subsection{Hierarchical Clustering}
We compare the DTW and OTW distance for time series clustering. We use the Agglomerative Clustering algorithm \cite{Ackermann2012AnalysisOA}, which only requires access to the distance matrix. We run such clustering algorithm on the UCR time series benchmark collection. We skip datasets having more than 500 samples, as the quadratic memory requirements of the algorithms imposes such constraint. We evaluate the quality of the clustering using the Rand Index (RI), given that the datasets in the UCR archive are labelled. We summarize the results in table \ref{tab:clustering-comparison}. We observe that our proposed distance outperforms DTW on most datasets considered. This is striking, as the time required for our method to run is considerably less than the DTW-based clustering. Overall, we observe in \cref{tab:clustering-comparison} that \textbf{OTW improves over DTW on 7 out of 7 types of datasets}. On some datasets like Image, Traffic and Sensor, the advantage is apparent.

\begin{table}[h]
\centering
\scriptsize
\caption{
            Average Rand Index (RI) and average cpu time (seconds) of OT-based
            and DTW-based Hierarchical Clustering methods.
            }
\label{tab:clustering-comparison}
\begin{tabular}{cccccc}
\toprule
\textbf{Type} & \textbf{DTW RI - (time)}  &   \textbf{OTW RI - (time)}  &  \textbf{Improv.} &  \textbf{Total} & \textbf{\%} \\
\midrule
       Device &          $0.40$        - $(2,279)$ &  $\mathbf{0.49}$ -  $\mathbf{(81)}$ &                     5 &               6 &    $\mathbf{83}$ \\
        Image &          $0.62$     -      $(277)$ &  $\mathbf{0.67}$ -  $\mathbf{(28)}$ &                    25 &              29 &    $\mathbf{86}$ \\
 Medical &          $0.79$      -     $(884)$ &  $\mathbf{0.82}$  - $\mathbf{(40)}$ &                     7 &               9 &    $\mathbf{78}$ \\
        Power &          $0.50$     -       $(90)$ &  $\mathbf{0.52}$ - $\mathbf{(20)}$ &                     1 &               1 &   $\mathbf{100}$ \\
       Sensor &          $0.57$     -      $(384)$ &  $\mathbf{0.67}$  - $\mathbf{(31)}$ &                    15 &              17 &    $\mathbf{88}$ \\
      Spectro &          $0.56$     -       $(88)$ &  $\mathbf{0.61}$  - $\mathbf{(11)}$ &                     4 &               7 &    $\mathbf{57}$ \\
      Traffic &          $0.62$    -        $(61)$ &  $\mathbf{0.72}$  -  $\mathbf{(9)}$ &                     2 &               2 &   $\mathbf{100}$ \\
\bottomrule
\end{tabular}
\end{table}

\subsection{Performance of DTW-Net vs OTW-Net in synthetic and real data}
DTW distances have been employed to design neural network layers with inductive biases that are better
suited for time series data. This is the case, for example, of DTW-Net \cite{Cai2019}. In this
network, the first hidden layer consists of DTW distances between the input and the rows of a matrix, which is the trainable parameter of the layer. If there are $k$ such rows, then the computational complexity of the layer is $O(kn^2)$, where $n$ is the length of the input. On top of such features an arbitrary network architecture is added, which outputs the class probabilities.

In contrast, in a vanilla multi-layer fully-connected neural network the first hidden-layer (a linear layer) consists of inner products between the input and the rows of a matrix. If there are $k$ such rows, the complexity of this linear layer is $O(kn)$. Hence, the complexity of DTW-Net \cite{Cai2019} is higher than a regular fully-connected neural network: it suffers from a computational bottleneck. In this experiment, we replace the DTW distance in DTW-Net by our OTW distance. Because OTW can be computed in linear time, this restores the complexity to $O(kn)$, in line with the other layers of the network, getting rid of the computational bottleneck. We describe such layers in \cref{alg:otw-layer,alg:dtw-layer}.

\begin{algorithm}[H]
	\caption{OTW Feature Extraction Layer}
	\label{alg:otw-layer}
	\begin{algorithmic}
	    \State \textbf{Input: } input $a \in \R^n$
		\State \textbf{Parameters:} matrix $B \in \R^{k \times n}$
		\For{$i=1$ to $k$}
		\State $b \gets B_{[:, i]}$ \Comment{$i$-th row of $B$}
		\State $z_i \gets \text{OTW}^\beta_{m, s}(a, b)$
		
		\EndFor
		\State \textbf{return } $z \in \mathbb{R}^k$
	\end{algorithmic}
\end{algorithm}

\begin{algorithm}[H]
	\caption{DTW Feature Extraction Layer}
	\label{alg:dtw-layer}
	\begin{algorithmic}
	    \State \textbf{Input: } $a \in \R^d$
		\State \textbf{Parameters:}  matrix $B \in \R^{k \times n}$
		\For{$i=1$ to $k$}
		\State $b \gets B_{[:, i]}$ \Comment{$i$-th row of $B$}
		\State $z_i \gets \text{DTW}(a, b)$
		
		\EndFor
		\State \textbf{return } $z \in \mathbb{R}^k$
	\end{algorithmic}
\end{algorithm}

\begin{figure*}
\centering
\includegraphics[width=0.32\textwidth]{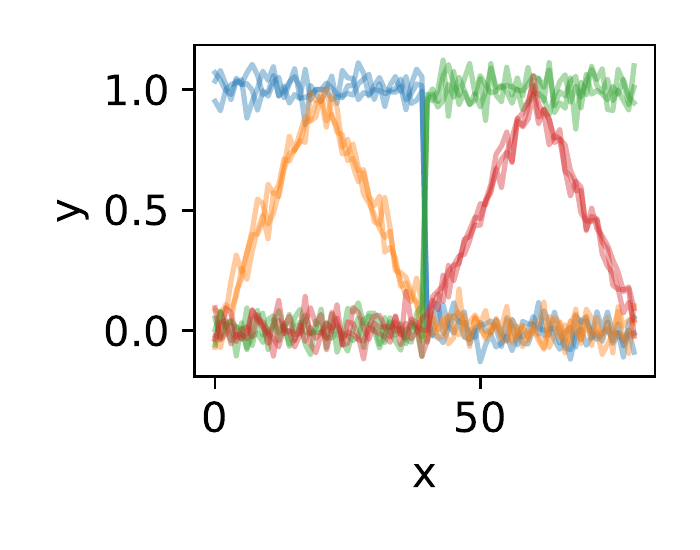}
\includegraphics[width=0.32\textwidth]{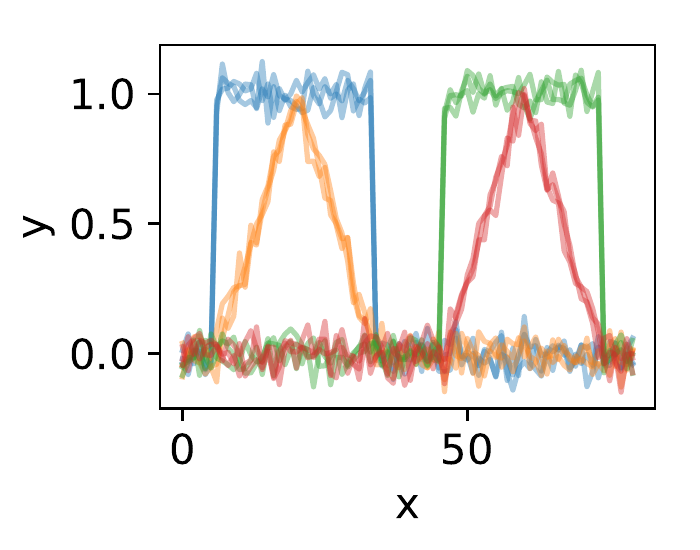}
\includegraphics[width=0.32\textwidth]{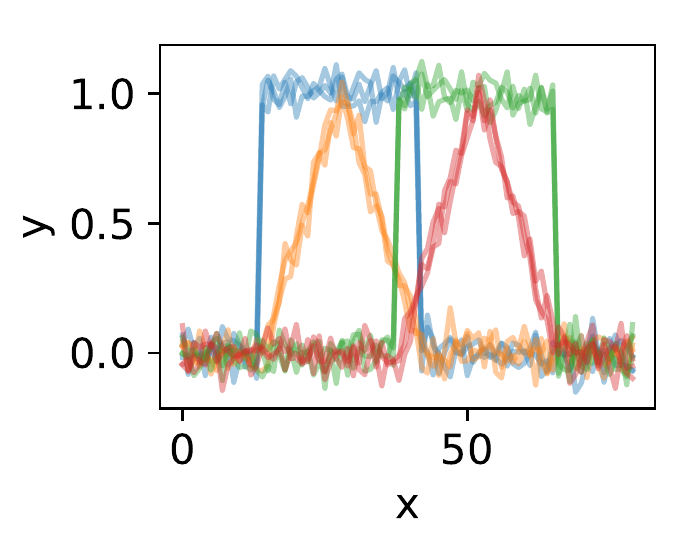}
\caption{The three synthetic labeled datasets considered, consisting of 4 different classes determined by a combination of shape (square/triangle) and time shift. Each color corresponds to a different class. 4 samples of each class are shown.}
\label{fig:synth_datasets}
\end{figure*}

\textbf{Synthetic data. } In order to illustrate the performance of both types of networks, we generate three synthetic datasets c.f. Figure \ref{fig:synth_datasets}. The synthetic datasets contain four classes which are determined by shape (triangle or square), location (left or right) and some added noise.  

\begin{figure*}
\centering
\includegraphics[width=0.99\textwidth]{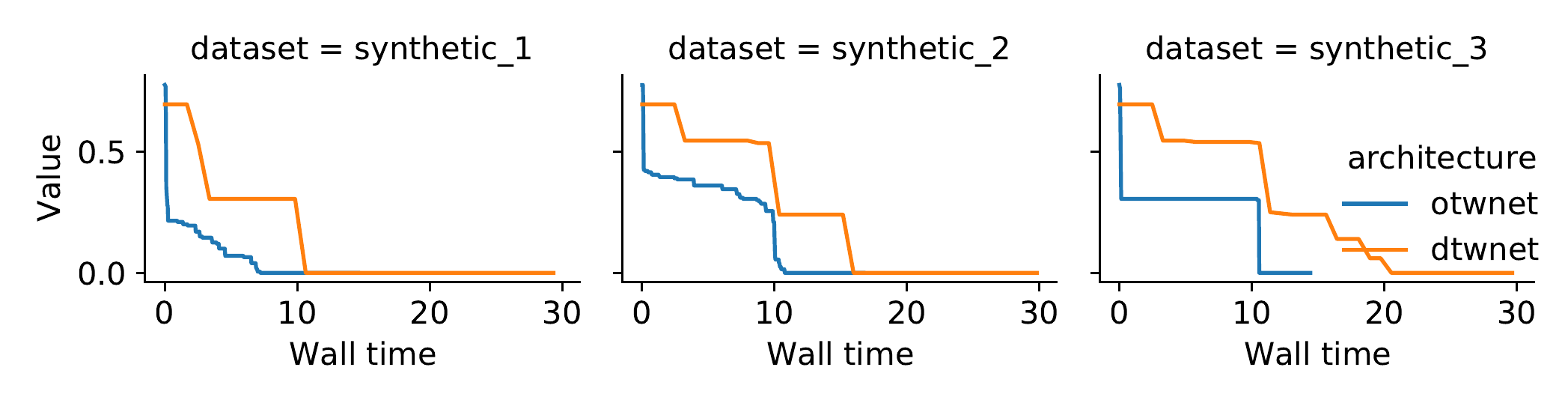}
\caption{Test error vs Wall clock time (in seconds) of DTW-Net and OTW-Net, trained on the synthetic datasets proposed.}
\label{fig:test_otnet_dtwnet}
\end{figure*}

\begin{figure*}
\centering
\includegraphics[width=0.32\textwidth]{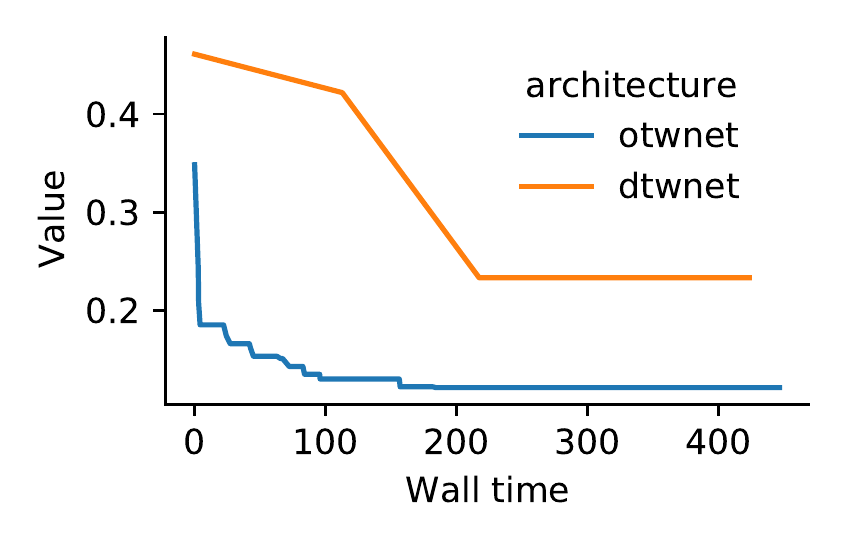}
\includegraphics[width=0.32\textwidth]{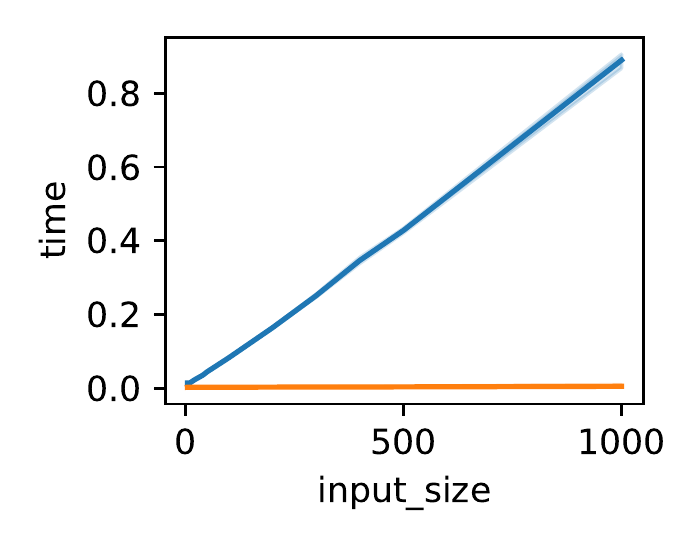}
\includegraphics[width=0.32\textwidth]{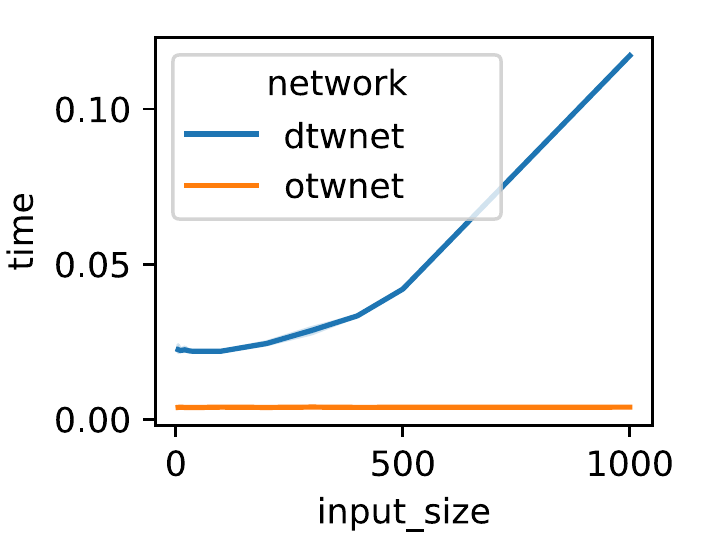}
\caption{Left: Test error vs Wall clock time of DTW-Net and OTW-Net, trained on the MoteStrain dataset from the UCR time series archive. Center: Wall clock time of a forward/backward pass over the network (in CPU) as a function of the size of the input. Right:  Wall clock time of a forward/backward pass over the network (in GPU) as a function of the size of the input.}
\label{fig:test_otnet_dtwnet_real}
\end{figure*}

For the synthetic data experiment, the hidden layer sizes for both DTW-Net and OTW-Net are set as $[1, 128, 128]$. We train both networks for 500 epochs and we plot the test-error vs training time in Figure \ref{fig:test_otnet_dtwnet}. To improve the readability, we plot the minimum achieved test error, up to time $t$. Due to the computational bottleneck in DTW-net, its training time is orders of magnitude larger than OTW-Net. However, DTW-Net converges in fewer epochs, which somewhat offsets its slower time-per-epoch. In any case, OTW-Net achieves zero-error in 50 to 60 percent of the time of DTW-Net. Note that even if the training time to convergence is only reduced by 50\%-60\%, \textbf{the inference is reduced by a larger margin} and is many times faster (see \cref{fig:test_otnet_dtwnet_real} middle and right panes), which shows the true advantage of linear complexity + GPU usage.

\textbf{Real data.} We compare DTW-Net with OTW-Net on real datasets from the UCR time series archive. For the real data experiment, the hidden layer sizes for OTW-Net are set as $[500, 500, 500]$ and for DTW-Net as $[100, 500, 500]$. The smaller size of the first hidden layer of DTW-Net allows training in a reasonable amount of time. In fact, we estimate the time and memory required to train DTW-Net on the UCR time series archive, consisting of more than 100 datasets, and we conclude that DTW-Net can only be trained on a handful of them in less than 24 hours. We choose the MoteStrain dataset in this collection to demonstrate the speed and accuracy of both methods.

We train both networks for 5000 epochs and we plot the test-error vs training time in Figure \ref{fig:test_otnet_dtwnet_real}. To improve the readability, we plot the minimum achieved test error, up to time $t$. For the MoteStrain dataset, we only show the time up to 400 seconds, when OTW-Net converges to around 14\% test error. In contrast, DTW-Net only achieves 25\% error and its training takes around 12 hours, compared to OTW-Net which only takes 30 minutes. We conclude that the quadratic complexity of the first layer in DTW-Net makes this approach unfeasible on realistic datasets, and that one way to solve this problem is to use our proposed architecture OTW-Net.

\textbf{Speed comparison on CPU/GPU. } We turn to understanding how the higher complexity of DTW affects the performance of Deep Neural Networks with DTW-based layers and our proposed OTW-based layer replacement. To this end, we compute the time that it takes to perform one forward/backward pass through the networks, for an input of increasing dimension.

In CPU, we compare our OTW-Net against DTW-Net. DTW does not have a GPU implementation available, so instead we use the state-of-the-art GPU implementation of Soft-DTW \cite{Maghoumi2020}. The results are plotted in Figure \ref{fig:test_otnet_dtwnet_real}, in the middle and right panes. As expected we observe a stark difference in time, and our OTW-Net runs considerably faster than the DTW-based counterparts, both in CPU and GPU. This illustrates why the DTW based layers are not a feasible alternative already for moderate dimensions and faster alternatives, like our proposed method, have to be developed if practitioners are to adopt such kind of architectures.

\section{Conclusion}
\label{sec:conclusion}
We have introduced OTW, a distance for time-series that overcomes the main limitations of DTW while retaining or improving its performance in downstream tasks. In this way, it opens the path to the use of time-series distances inside Deep Learning pipelines, thanks to its easy GPU implementation and the absence of computational bottlenecks. One of the limitations of our work is the constraint that the time-series be one-dimensional. We leave possible extensions to the multivariate case as a promising direction of future research. Finally, we recall that despite the improvements, the purpose is not to completely replace DTW: for many datasets it is still the best performing distance. Rather, OTW is a complementary distance that should be preferred when there are resource constraints.

%{\color{red} {General short summary and contributions - mention any key limitations, future work - do we say things like DTW can work for arbitrary length, and we cannot? extension to multivariate?}}

\section*{Acknowledgements}
Fabian Latorre would like to thank Jorge Barreras, Chris Russel, Gregory Durrett, Laura Montoya, Lucia Cipolina-Kun and two  anonymous donors for their financial support\footnote{\url{https://www.gofundme.com/f/help-me-attend-icassp-2023}}. Without their help, it would have been impossible to present this work at ICASSP 2023.

\bibliographystyle{IEEEbib}
\bibliography{references}

\onecolumn
\newpage
\appendix

\section{1-Nearest-Neighbors Comparison}
\label{sec:nn-full-comparison}

\begin{table}[h]
\centering
\caption{
            Test error of 1-NN classification using Learned
            DTW or Learned OT.
            }
\label{tab:1nn-rank-comparison-full}
\small
\begin{tabular}{cccc}
\toprule
                  \textbf{Name} & \textbf{Type} & \textbf{DTW error} &         \textbf{OT error} \\
\midrule
                          ACSF1 &        Device &               0.38 &  $\mathbf{0.34 \pm 0.06}$ \\
                          Adiac &         Image &               0.39 &  $\mathbf{0.39 \pm 0.01}$ \\
             AllGestureWiimoteX &        Sensor &               0.28 &           $0.90 \pm 0.00$ \\
             AllGestureWiimoteY &        Sensor &               0.27 &           $0.90 \pm 0.00$ \\
             AllGestureWiimoteZ &        Sensor &               0.35 &           $0.90 \pm 0.00$ \\
                      ArrowHead &         Image &               0.20 &  $\mathbf{0.23 \pm 0.03}$ \\
                           Beef &       Spectro &               0.33 &           $0.55 \pm 0.04$ \\
                      BeetleFly &         Image &               0.30 &  $\mathbf{0.26 \pm 0.06}$ \\
                    BirdChicken &         Image &               0.30 &  $\mathbf{0.29 \pm 0.08}$ \\
                            Car &        Sensor &               0.23 &  $\mathbf{0.27 \pm 0.06}$ \\
                      Chinatown &       Traffic &               0.05 &  $\mathbf{0.07 \pm 0.03}$ \\
          ChlorineConcentration &        Sensor &               0.35 &           $0.39 \pm 0.02$ \\
                         Coffee &       Spectro &               0.00 &           $0.04 \pm 0.03$ \\
                      Computers &        Device &               0.38 &  $\mathbf{0.40 \pm 0.02}$ \\
                           Crop &         Image &               0.29 &  $\mathbf{0.28 \pm 0.01}$ \\
            DiatomSizeReduction &         Image &               0.07 &  $\mathbf{0.06 \pm 0.03}$ \\
   DistalPhalanxOutlineAgeGroup &         Image &               0.37 &  $\mathbf{0.23 \pm 0.03}$ \\
    DistalPhalanxOutlineCorrect &         Image &               0.28 &  $\mathbf{0.23 \pm 0.01}$ \\
                DistalPhalanxTW &         Image &               0.37 &  $\mathbf{0.29 \pm 0.04}$ \\
                  DodgerLoopDay &        Sensor &               0.41 &           $0.85 \pm 0.03$ \\
                 DodgerLoopGame &        Sensor &               0.07 &           $0.43 \pm 0.09$ \\
              DodgerLoopWeekend &        Sensor &               0.02 &           $0.35 \pm 0.09$ \\
                         ECG200 &  Medical Data &               0.12 &           $0.16 \pm 0.04$ \\
                        ECG5000 &  Medical Data &               0.07 &  $\mathbf{0.08 \pm 0.01}$ \\
                    ECGFiveDays &  Medical Data &               0.20 &  $\mathbf{0.19 \pm 0.03}$ \\
            EOGHorizontalSignal &  Medical Data &               0.52 &  $\mathbf{0.38 \pm 0.02}$ \\
              EOGVerticalSignal &  Medical Data &               0.52 &  $\mathbf{0.49 \pm 0.01}$ \\
                    Earthquakes &        Sensor &               0.27 &           $0.32 \pm 0.03$ \\
                ElectricDevices &        Device &               0.38 &  $\mathbf{0.25 \pm 0.00}$ \\
                   EthanolLevel &       Spectro &               0.72 &  $\mathbf{0.71 \pm 0.01}$ \\
                        FaceAll &         Image &               0.19 &  $\mathbf{0.05 \pm 0.01}$ \\
                       FaceFour &         Image &               0.11 &           $0.17 \pm 0.05$ \\
                       FacesUCR &         Image &               0.09 &           $0.11 \pm 0.01$ \\
                     FiftyWords &         Image &               0.24 &           $0.31 \pm 0.01$ \\
                           Fish &         Image &               0.15 &           $0.21 \pm 0.02$ \\
                          FordA &        Sensor &               0.31 &  $\mathbf{0.27 \pm 0.01}$ \\
                          FordB &        Sensor &               0.39 &  $\mathbf{0.30 \pm 0.01}$ \\
            FreezerRegularTrain &        Sensor &               0.09 &  $\mathbf{0.08 \pm 0.02}$ \\
              FreezerSmallTrain &        Sensor &               0.32 &  $\mathbf{0.24 \pm 0.02}$ \\

\bottomrule
\end{tabular}
\end{table}

\begin{table}
\centering
\caption{
            Test error of 1-NN classification using Learned
            DTW or Learned OT (continued).
            }
\label{tab:1nn-rank-comparison-full}
\small \begin{tabular}{cccc}
\toprule
                  \textbf{Name} & \textbf{Type} & \textbf{DTW error} &         \textbf{OT error} \\
\midrule
                          Fungi &  Medical Data &               0.18 &           $0.41 \pm 0.05$ \\
                GesturePebbleZ1 &        Sensor &               0.17 &           $0.84 \pm 0.01$ \\
                GesturePebbleZ2 &        Sensor &               0.22 &           $0.83 \pm 0.02$ \\
                            Ham &       Spectro &               0.40 &  $\mathbf{0.28 \pm 0.03}$ \\
                   HandOutlines &         Image &               0.14 &  $\mathbf{0.15 \pm 0.01}$ \\
                        Herring &         Image &               0.47 &  $\mathbf{0.49 \pm 0.04}$ \\
                    HouseTwenty &        Device &               0.06 &           $0.25 \pm 0.04$ \\
          InsectEPGRegularTrain &  Medical Data &               0.17 &  $\mathbf{0.00 \pm 0.00}$ \\
            InsectEPGSmallTrain &  Medical Data &               0.31 &  $\mathbf{0.00 \pm 0.00}$ \\
            InsectWingbeatSound &        Sensor &               0.42 &           $0.45 \pm 0.01$ \\
               ItalyPowerDemand &        Sensor &               0.04 &           $0.07 \pm 0.02$ \\
         LargeKitchenAppliances &        Device &               0.21 &           $0.41 \pm 0.01$ \\
                     Lightning2 &        Sensor &               0.13 &           $0.22 \pm 0.05$ \\
                     Lightning7 &        Sensor &               0.29 &           $0.35 \pm 0.04$ \\
                           Meat &       Spectro &               0.07 &  $\mathbf{0.01 \pm 0.01}$ \\
                  MedicalImages &         Image &               0.25 &           $0.29 \pm 0.02$ \\
            MelbournePedestrian &       Traffic &               0.18 &  $\mathbf{0.05 \pm 0.00}$ \\
   MiddlePhalanxOutlineAgeGroup &         Image &               0.48 &  $\mathbf{0.31 \pm 0.03}$ \\
    MiddlePhalanxOutlineCorrect &         Image &               0.23 &  $\mathbf{0.22 \pm 0.01}$ \\
                MiddlePhalanxTW &         Image &               0.49 &  $\mathbf{0.45 \pm 0.03}$ \\
          MixedShapesSmallTrain &         Image &               0.17 &  $\mathbf{0.19 \pm 0.02}$ \\
                     MoteStrain &        Sensor &               0.13 &  $\mathbf{0.15 \pm 0.03}$ \\
     NonInvasiveFetalECGThorax1 &  Medical Data &               0.19 &  $\mathbf{0.18 \pm 0.01}$ \\
                        OSULeaf &         Image &               0.39 &  $\mathbf{0.42 \pm 0.04}$ \\
                       OliveOil &       Spectro &               0.13 &  $\mathbf{0.14 \pm 0.03}$ \\
                          PLAID &        Device &               0.17 &           $0.88 \pm 0.04$ \\
       PhalangesOutlinesCorrect &         Image &               0.24 &  $\mathbf{0.23 \pm 0.01}$ \\
                        Phoneme &        Sensor &               0.77 &           $0.88 \pm 0.01$ \\
          PickupGestureWiimoteZ &        Sensor &               0.34 &           $0.90 \pm 0.02$ \\
              PigAirwayPressure &  Medical Data &               0.90 &  $\mathbf{0.85 \pm 0.01}$ \\
                 PigArtPressure &  Medical Data &               0.80 &  $\mathbf{0.78 \pm 0.02}$ \\
                         PigCVP &  Medical Data &               0.84 &           $0.86 \pm 0.02$ \\
                          Plane &        Sensor &               0.00 &           $0.05 \pm 0.01$ \\
                      PowerCons &         Power &               0.08 &  $\mathbf{0.04 \pm 0.01}$ \\
 ProximalPhalanxOutlineAgeGroup &         Image &               0.21 &           $0.24 \pm 0.02$ \\
  ProximalPhalanxOutlineCorrect &         Image &               0.21 &  $\mathbf{0.21 \pm 0.02}$ \\
              ProximalPhalanxTW &         Image &               0.24 &           $0.27 \pm 0.02$ \\
           RefrigerationDevices &        Device &               0.56 &  $\mathbf{0.49 \pm 0.02}$ \\
                     ScreenType &        Device &               0.59 &  $\mathbf{0.57 \pm 0.01}$ \\
           ShakeGestureWiimoteZ &        Sensor &               0.16 &           $0.90 \pm 0.01$ \\
                      ShapesAll &         Image &               0.20 &           $0.27 \pm 0.02$ \\
         SmallKitchenAppliances &        Device &               0.33 &           $0.37 \pm 0.02$ \\
          SonyAIBORobotSurface1 &        Sensor &               0.30 &  $\mathbf{0.14 \pm 0.03}$ \\
          SonyAIBORobotSurface2 &        Sensor &               0.14 &  $\mathbf{0.14 \pm 0.02}$ \\
                     Strawberry &       Spectro &               0.05 &  $\mathbf{0.04 \pm 0.01}$ \\
                    SwedishLeaf &         Image &               0.15 &           $0.23 \pm 0.01$ \\
                        Symbols &         Image &               0.06 &           $0.14 \pm 0.03$ \\
                          Trace &        Sensor &               0.01 &           $0.21 \pm 0.03$ \\
                     TwoLeadECG &  Medical Data &               0.13 &           $0.31 \pm 0.06$ \\
                          Wafer &        Sensor &               0.00 &  $\mathbf{0.01 \pm 0.00}$ \\
                           Wine &       Spectro &               0.39 &  $\mathbf{0.13 \pm 0.07}$ \\
                   WordSynonyms &         Image &               0.26 &           $0.36 \pm 0.02$ \\
                           Yoga &         Image &               0.16 &           $0.17 \pm 0.01$ \\
\bottomrule
\end{tabular}
\end{table}

\newpage 
\section{Hierarchical Clustering Comparison}
\label{sec:clustering-full-comparison}
\begin{table}[h]
\centering
\caption{
            RI and time taken for DTW- and OT-based Hierarchical clustering
            methods on datasets in the UCR time-series benchmark collection.
            Noticeable improvements in boldface.
            }
\label{tab:clustering-comparison}
\tiny \begin{tabular}{cccccc}
\toprule
                  \textbf{Name} & \textbf{Type} & \textbf{DTW RI} & \textbf{DTW time} &   \textbf{OT RI} &  \textbf{OT time} \\
\midrule
                          Adiac &         Image &          $0.79$ &          $220.82$ &           $0.61$ &  $\mathbf{32.83}$ \\
                      ArrowHead &         Image &          $0.35$ &           $63.86$ &  $\mathbf{0.36}$ &  $\mathbf{14.33}$ \\
                           Beef &       Spectro &          $0.42$ &           $14.85$ &  $\mathbf{0.60}$ &   $\mathbf{2.56}$ \\
                      BeetleFly &         Image &          $0.59$ &           $12.97$ &  $\mathbf{0.62}$ &   $\mathbf{1.68}$ \\
                    BirdChicken &         Image &          $0.50$ &           $13.07$ &  $\mathbf{0.51}$ &   $\mathbf{1.93}$ \\
                            Car &        Sensor &          $0.48$ &           $94.91$ &  $\mathbf{0.66}$ &  $\mathbf{11.69}$ \\
                      Chinatown &       Traffic &          $0.59$ &           $42.02$ &  $\mathbf{0.61}$ &   $\mathbf{5.93}$ \\
          ChlorineConcentration &        Sensor &          $0.42$ &          $200.71$ &  $\mathbf{0.50}$ &  $\mathbf{32.12}$ \\
                         Coffee &       Spectro &          $0.50$ &            $5.59$ &  $\mathbf{0.57}$ &   $\mathbf{1.64}$ \\
                      Computers &        Device &          $0.50$ &        $2,495.25$ &  $\mathbf{0.50}$ &  $\mathbf{94.26}$ \\
                           Crop &         Image &          $0.87$ &           $90.27$ &  $\mathbf{0.92}$ &  $\mathbf{17.05}$ \\
            DiatomSizeReduction &         Image &          $0.30$ &          $250.90$ &  $\mathbf{0.31}$ &  $\mathbf{31.80}$ \\
   DistalPhalanxOutlineAgeGroup &         Image &          $0.71$ &          $103.39$ &  $\mathbf{0.71}$ &  $\mathbf{22.53}$ \\
    DistalPhalanxOutlineCorrect &         Image &          $0.53$ &          $103.47$ &  $\mathbf{0.53}$ &  $\mathbf{22.43}$ \\
                DistalPhalanxTW &         Image &          $0.79$ &          $102.95$ &  $\mathbf{0.87}$ &  $\mathbf{22.32}$ \\
                         ECG200 &  Medical Data &          $0.54$ &           $18.45$ &  $\mathbf{0.61}$ &   $\mathbf{5.63}$ \\
                        ECG5000 &  Medical Data &          $0.87$ &          $161.50$ &  $\mathbf{0.89}$ &  $\mathbf{27.41}$ \\
                    ECGFiveDays &  Medical Data &          $0.50$ &          $157.94$ &  $\mathbf{0.51}$ &  $\mathbf{26.80}$ \\
                    Earthquakes &        Sensor &          $0.51$ &        $1,594.13$ &  $\mathbf{0.68}$ &  $\mathbf{59.17}$ \\
                ElectricDevices &        Device &          $0.56$ &          $114.64$ &           $0.53$ &  $\mathbf{23.78}$ \\
                        FaceAll &         Image &          $0.57$ &          $147.43$ &  $\mathbf{0.70}$ &  $\mathbf{27.11}$ \\
                       FaceFour &         Image &          $0.54$ &           $30.19$ &  $\mathbf{0.75}$ &   $\mathbf{6.03}$ \\
                       FacesUCR &         Image &          $0.57$ &          $149.10$ &  $\mathbf{0.78}$ &  $\mathbf{27.17}$ \\
                     FiftyWords &         Image &          $0.92$ &          $400.92$ &  $\mathbf{0.95}$ &  $\mathbf{41.76}$ \\
                           Fish &         Image &          $0.17$ &          $486.88$ &  $\mathbf{0.64}$ &  $\mathbf{39.02}$ \\
                          FordA &        Sensor &          $0.50$ &        $1,220.33$ &  $\mathbf{0.50}$ &  $\mathbf{66.80}$ \\
                          FordB &        Sensor &          $0.51$ &        $1,222.25$ &  $\mathbf{0.51}$ &  $\mathbf{66.06}$ \\
            FreezerRegularTrain &        Sensor &          $0.51$ &          $474.91$ &  $\mathbf{0.56}$ &  $\mathbf{46.98}$ \\
              FreezerSmallTrain &        Sensor &          $0.50$ &          $480.85$ &  $\mathbf{0.66}$ &  $\mathbf{48.18}$ \\
                          Fungi &  Medical Data &          $0.96$ &           $43.36$ &  $\mathbf{0.98}$ &  $\mathbf{12.53}$ \\
                            Ham &       Spectro &          $0.50$ &          $161.65$ &  $\mathbf{0.50}$ &  $\mathbf{19.27}$ \\
                        Herring &         Image &          $0.51$ &          $121.49$ &  $\mathbf{0.52}$ &  $\mathbf{11.87}$ \\
          InsectEPGRegularTrain &  Medical Data &          $1.00$ &          $669.58$ &  $\mathbf{1.00}$ &  $\mathbf{37.38}$ \\
            InsectEPGSmallTrain &  Medical Data &          $1.00$ &          $625.76$ &  $\mathbf{1.00}$ &  $\mathbf{32.23}$ \\
            InsectWingbeatSound &        Sensor &          $0.68$ &          $426.09$ &  $\mathbf{0.85}$ &  $\mathbf{40.46}$ \\
               ItalyPowerDemand &        Sensor &          $0.50$ &           $78.79$ &  $\mathbf{0.50}$ &  $\mathbf{12.79}$ \\
         LargeKitchenAppliances &        Device &          $0.34$ &        $2,817.32$ &  $\mathbf{0.57}$ &  $\mathbf{91.34}$ \\
                     Lightning2 &        Sensor &          $0.50$ &          $114.12$ &  $\mathbf{0.61}$ &  $\mathbf{12.99}$ \\
                     Lightning7 &        Sensor &          $0.62$ &           $43.17$ &  $\mathbf{0.82}$ &   $\mathbf{8.52}$ \\
                           Meat &       Spectro &          $0.77$ &           $62.18$ &  $\mathbf{0.77}$ &  $\mathbf{10.92}$ \\
                  MedicalImages &         Image &          $0.60$ &          $119.65$ &           $0.57$ &  $\mathbf{24.08}$ \\
            MelbournePedestrian &       Traffic &          $0.64$ &           $79.33$ &  $\mathbf{0.83}$ &  $\mathbf{12.70}$ \\
   MiddlePhalanxOutlineAgeGroup &         Image &          $0.70$ &          $103.33$ &  $\mathbf{0.70}$ &  $\mathbf{22.24}$ \\
    MiddlePhalanxOutlineCorrect &         Image &          $0.52$ &          $103.52$ &           $0.52$ &  $\mathbf{22.27}$ \\
                MiddlePhalanxTW &         Image &          $0.80$ &          $102.36$ &  $\mathbf{0.81}$ &  $\mathbf{22.51}$ \\
                     MoteStrain &        Sensor &          $0.50$ &          $105.35$ &  $\mathbf{0.68}$ &  $\mathbf{22.53}$ \\
     NonInvasiveFetalECGThorax1 &  Medical Data &          $0.83$ &        $3,068.55$ &  $\mathbf{0.93}$ &  $\mathbf{95.15}$ \\
     NonInvasiveFetalECGThorax2 &  Medical Data &          $0.92$ &        $3,112.35$ &  $\mathbf{0.94}$ &  $\mathbf{94.65}$ \\
                        OSULeaf &         Image &          $0.58$ &          $700.76$ &  $\mathbf{0.69}$ &  $\mathbf{49.36}$ \\
                       OliveOil &       Spectro &          $0.74$ &           $27.54$ &  $\mathbf{0.78}$ &   $\mathbf{2.93}$ \\
       PhalangesOutlinesCorrect &         Image &          $0.53$ &          $108.40$ &  $\mathbf{0.53}$ &  $\mathbf{22.23}$ \\
                          Plane &        Sensor &          $0.96$ &           $30.50$ &           $0.91$ &   $\mathbf{8.11}$ \\
                      PowerCons &         Power &          $0.50$ &           $89.67$ &  $\mathbf{0.52}$ &  $\mathbf{19.95}$ \\
 ProximalPhalanxOutlineAgeGroup &         Image &          $0.77$ &          $104.00$ &  $\mathbf{0.78}$ &  $\mathbf{22.52}$ \\
  ProximalPhalanxOutlineCorrect &         Image &          $0.53$ &          $107.66$ &  $\mathbf{0.56}$ &  $\mathbf{22.33}$ \\
              ProximalPhalanxTW &         Image &          $0.85$ &          $107.30$ &  $\mathbf{0.87}$ &  $\mathbf{22.04}$ \\
           RefrigerationDevices &        Device &          $0.34$ &        $2,757.08$ &  $\mathbf{0.44}$ &  $\mathbf{91.89}$ \\
                     ScreenType &        Device &          $0.34$ &        $2,845.35$ &  $\mathbf{0.52}$ &  $\mathbf{91.20}$ \\
                      ShapesAll &         Image &          $0.83$ &        $1,979.40$ &  $\mathbf{0.94}$ &  $\mathbf{67.31}$ \\
         SmallKitchenAppliances &        Device &          $0.34$ &        $2,646.25$ &  $\mathbf{0.37}$ &  $\mathbf{91.76}$ \\
          SonyAIBORobotSurface1 &        Sensor &          $0.50$ &           $96.16$ &  $\mathbf{0.75}$ &  $\mathbf{21.52}$ \\
          SonyAIBORobotSurface2 &        Sensor &          $0.52$ &           $93.90$ &  $\mathbf{0.67}$ &  $\mathbf{21.03}$ \\
                     Strawberry &       Spectro &          $0.52$ &          $327.31$ &  $\mathbf{0.54}$ &  $\mathbf{38.26}$ \\
                    SwedishLeaf &         Image &          $0.37$ &          $167.14$ &  $\mathbf{0.52}$ &  $\mathbf{26.56}$ \\
                        Symbols &         Image &          $0.87$ &          $778.76$ &  $\mathbf{0.87}$ &  $\mathbf{60.72}$ \\
                          Trace &        Sensor &          $0.87$ &           $65.96$ &           $0.76$ &  $\mathbf{14.12}$ \\
                     TwoLeadECG &  Medical Data &          $0.50$ &          $104.65$ &  $\mathbf{0.51}$ &  $\mathbf{22.13}$ \\
                          Wafer &        Sensor &          $0.53$ &          $183.66$ &  $\mathbf{0.80}$ &  $\mathbf{29.94}$ \\
                           Wine &       Spectro &          $0.50$ &           $16.15$ &  $\mathbf{0.50}$ &   $\mathbf{3.76}$ \\
                   WordSynonyms &         Image &          $0.85$ &          $407.55$ &  $\mathbf{0.89}$ &  $\mathbf{41.29}$ \\
                           Yoga &         Image &          $0.50$ &          $865.61$ &  $\mathbf{0.51}$ &  $\mathbf{58.34}$ \\
                          Total &             - &          $0.60$ &          $521.74$ &  $\mathbf{0.67}$ &  $\mathbf{32.15}$ \\
\bottomrule
\end{tabular}
\end{table}

\newpage
\section{Deferred Proofs}
\label{sec:proof_thm_1}
\subsection{Proof of \cref{thm:unbalanced-solution}}
First we will prove the following intermediate result:
\begin{lemma} \label{lem:aux1}
Let $a,b$ be two positive sequences. Without loss of generality assume $\sum_{j=1}^n a_j \leq \sum_{j=1}^n b_j$ and let $A(i)=\sum_{j=1}^i a_j, B(i)=\sum_{j=1}^i b_j$ be the partial sums of the sequences. It holds that:
\begin{equation}
    \widehat{W}_m(a, b) \leq  \begin{cases} \min_c W(a, c) + m |A(n) - B(n)| \\
    \text{subject to }  0 \leq c \leq b, \sum_{j=1}^n c_j = \sum_{j=1}^n a_j
    \end{cases}
\end{equation}
Where $W(a, c)$ is the original (\cref{eq:ot}) Optimal Transport distance (note that $a, b$ sum up to the same value).
\end{lemma}
\begin{proof}
Let $c$ such that $0 \leq c 
\leq b$ and $\sum_{j=1}^n c_j = \sum_{j=1}^n a_j$,
Let $T^\star$ be the optimal transport map between $a$ and $c$. That means $T^\star \geq 0, T^\star \mathbbm{1} = a, \mathbbm{1}^\top T^\star = c^\top$. We now build a feasible transport map for the unbalanced problem \cref{eq:unbalanced-ot} based on $T^\star$:
\begin{equation}
    \widehat{T} = \left [
    \begin{array}{cc} T^\star & 0
    \\ b^\top - c^\top &  \sum_{j=1}^n a_i
    \end{array} \right ]
\end{equation}
then it is easy to see that $\widehat{T} \mathbbm{1} = \hat{a}$ and $\mathbbm{1}^\top \widehat{T} = \hat{b}$, so $\widehat{T}$ is a feasible transport map for \cref{eq:unbalanced-ot}. Computing the objective value we obtain:
\begin{equation}
\begin{split}
    \widehat{W}_m(a, b) &\leq \left \langle \widehat{T}, D(m) \right \rangle \\
    &= \langle T^\star, D \rangle + m \left (\sum_{j=1}^n b_j - \sum_{j=1}^n c_j \right ) \\
    &= \langle T^\star, D \rangle + m \left (\sum_{j=1}^n b_j - \sum_{j=1}^n a_j \right ) \\
    & = W(a, c) + m|A(n) - B(n)|
\end{split}
\end{equation}
\end{proof}
minimizing the right hand side we obtain:
\begin{equation}
    \widehat{W}_m(a, b) \leq  \begin{cases} \min_c W(a, c) + m |A(n) - B(n)| \\
    \text{subject to }  0 \leq c \leq b, \sum_{j=1}^n c_j = \sum_{j=1}^n a_j
    \end{cases}
\end{equation}
Now we will define a nonnegative sequence $c$ as follows: let $i^\star \in [n]$ be such that
\begin{equation}
    \sum_{j=1}^{i^\star} b_j \geq \sum_{j=1}^n a_j, 
 \qquad \sum_{j=1}^{i^\star -1}b_j < \sum_{j=1}^n a_j
\end{equation}
that is, $i^\star$ is exactly the index where the cumulative mass of $b$ exceeds the total mass of $a$. This index exists because without loss of generality we can assume $\sum_{j=1}^n a_j \leq \sum_{j=1}^n b_j$. Now we define
\begin{equation}
    c_i = \begin{cases} b_i &: \text{if } i=1, \ldots, i^\star - 1 \\
    \sum_{j=1}^n - \sum_{j=1}^{i^\star} b_j &: \text{if } i = i^\star \\
    0 &: \text{if } i > i^{\star}
    \end{cases}
\end{equation}
then it is easy to see that $\sum_{j=1}^n c_i=\sum_{j=1}^n a_i$. Then for this choice of $c$ it holds that 
\begin{equation}
\widehat{W}_m(a, b) \leq W(a, c) + m|A(n) - B(n)|
\end{equation}
Now we have a formula for $W(a, c)$ according to \cite{Vallender1974}:
\begin{equation}
    W(a, c) = \sum_{j=1}^n |A(j) - C(j)|, \qquad C(j) = \sum_{i=1}^j c_i
\end{equation}
Now we have
\begin{equation}
    \begin{split}
        \sum_{j=1}^n |A(j) - C(j)| & = \sum_{j=1}^{n-1}|A(j) - C(j)| \qquad (\text{because } A(n)=C(n))\\
        & = \sum_{j=1}^{i^\star - 1} |A(j) - C(j)| +
        \sum_{j=i^\star}^{n-1} C(j) - A(j) \qquad (\text{because } C(j) \geq A(j) \text{ for } j \geq i^\star) \\
        &\leq \sum_{j=1}^{i^\star - 1} |A(j) - C(j)|
        + \sum_{j=i^\star}^{n-1} B(j) - A(j) \qquad (\text{because } C(j) \leq B(j) \text{ for } j \geq i^\star) \\
        &= \sum_{j=1}^{i^\star - 1} |A(j) - B(j)|
        + \sum_{j=i^\star}^{n-1} B(j) - A(j) \qquad (\text{because } C(j) = B(j) \text{ for } j < i^\star) \\
        &= \sum_{j=1}^{n-1}|A(j) - B(j)|
    \end{split}
\end{equation}
Hence, by \cref{lem:aux1} we have
\begin{equation}
\widehat{W}_m(a, b) \leq \sum_{j=1}^{n-1}|A(j) - B(j)| + m|A(n) - B(n)|
\end{equation}

\subsection{Proof of \cref{lem:shift-sensitivity}}
Without loss of generality we again assume that $\sum_{i=1}^n a_i \leq \sum_{i=1}^n b_i$. We will show that
\begin{equation}
|\widehat{W}_m(a, b') - \widehat{W}_m(a, b)| \leq t \left (\sum_{i=1}^n a_i\right )
\end{equation}
To this end we will first show that
\begin{equation}
\label{eq:basic-ineq}
    \widehat{W}(a, b') \leq
    \widehat{W}(a, b) + t \left (
    \sum_{i=1}^n a_i \right )
\end{equation}
then by symmetry (we can swap the roles of $b$ and $b'$ and follow the same argument) we can conclude:
\begin{equation}
\label{eq:basic-ineq2}
    \widehat{W}(a, b) \leq
    \widehat{W}(a, b') + t \left (
    \sum_{i=1}^n a_i \right )
\end{equation}
so that
\begin{equation}
    \begin{split}
    \label{eq:basic-ineq3}
        |\widehat{W}_m(a, b') - \widehat{W}_m(a, b)| &=
        \max \left (\widehat{W}_m(a, b') - \widehat{W}_m(a, b), \widehat{W}_m(a, b) - \widehat{W}_m(a, b')\right ) \\
        & \leq t \left (
    \sum_{i=1}^n a_i \right )
    \end{split}
\end{equation}
Leading to the result. Going back to prove \cref{eq:basic-ineq}, let $T^\star$ be an unbalanced optimal transport map between $a$ and $b$. From this map, we will construct a transport map between $a, b'$ which has cost equal to the right hand side of \cref{eq:basic-ineq}. Recall that unbalanced transport maps (\cref{eq:unbalanced-ot}) can be split in the following block form
\begin{equation}
\label{eq:blocks}
    T^\star = \left [ \begin{array}{cc}
    T' & \tilde{a} \\
    \tilde{b}^\top & c
    \end{array} \right ]
\end{equation}
where $T$ has dimensions $n \times n$, $\tilde{a}$ is a column vector of length $n$, $\tilde{b}$ is a vector of length $n$ and $c$ is a scalar. This follows the structure of the matrix
\begin{equation}
    D(m) = \left [ \begin{array}{cc}
    D & m \mathbbm{1} \\
    m \mathbbm{1}^\top & 0
    \end{array} \right ]
\end{equation}
The fact that the column-sums of the transport map should be equal to $b$, (this is a constraint in the definition \cref{eq:unbalanced-ot}) which by assumption is such that $b_{n-t+1}=\ldots=b_n=0$ implies that the left-most block in \cref{eq:blocks} can be written as:
\begin{equation}
    \label{eq:blocks2}
    \left [\begin{array}{c}
    T' \\
    \tilde{b}^\top
    \end{array} \right ]
    = \left [\begin{array}{cc}
    T'' & \mathbf{0}_1 \\
    \tilde{b}'^\top & \mathbf{0}_2^\top
    \end{array} \right ]
\end{equation}
where $T''$ has dimensions $n \times (n-t)$, $\tilde{b'}$ is a vector of length $n-t$, $\mathbf{0}_1$ is a matrix of all-zeros with dimensions $n \times t$ and $\mathbf{0}_2$ is a zero vector of length $t$. The new transport map between $a$ and $b'$ is defined as
\begin{equation}
    \label{eq:blocks3}
    \widehat{T} = \left [ \begin{array}{cc}
    T''' & \tilde{a} \\
    \tilde{b''}^\top & c
    \end{array} \right ], \qquad
    \left [\begin{array}{c}
    T''' \\
    \tilde{b''}^\top
    \end{array} \right ]
    = \left [\begin{array}{cc}
    \mathbf{0}_1 & T'' \\
    \mathbf{0}_2 & \tilde{b}'^\top
    \end{array} \right ]
\end{equation}
that is, the new transport map is obtained by shifting to the right the left block in the decomposition \cref{eq:blocks} by $t$ units. The following observations hold: (1) after this transformation there is no change in the row-sums and (2) because $b'$ is obtained precisely by shifting $b$ to the right by $t$ units, the new unbalanced transport map satisfies 
\begin{equation}
\mathbbm{1}^\top \widehat{T} = \left [b' | 
 \sum_{i=1}^n a_i \right ] = \hat{b'}, \qquad
  \widehat{T} \mathbbm{1} = \left [a | 
 \sum_{i=1}^n b_i \right ] = \left [a | 
 \sum_{i=1}^n b'_i \right ] = \hat{a},
\end{equation}
Hence it is indeed the case that $\widehat{T}$ is an unbalanced transport map between $a$ and $b'$. We now compute the objective value:
\begin{equation}
\label{eq:combine1}
\begin{split}
    \widehat{W}(a, b') \leq \left \langle \widehat{T}, D(m) \right \rangle &= 
    \left \langle \left [ \begin{array}{cc}
    T''' & \tilde{a} \\
    \tilde{b''}^\top & c
    \end{array} \right ],  \left [ \begin{array}{cc}
    D & m \mathbbm{1} \\
    m \mathbbm{1}^\top & 0
    \end{array} \right ] \right \rangle \\
    &= \langle T''', D \rangle + m \sum_{i=1}^n \tilde{a}_i + m \sum_{i=1}^n \tilde{b''}_i \\
    &= \langle T''', D \rangle + m \sum_{i=1}^n \tilde{a}_i + m \sum_{i=1}^n \tilde{b}_i \qquad (\tilde{b''} \text{ is a shifted version of } \tilde{b}) 
    \end{split}
\end{equation}
We will now show the following:
\begin{equation}
\label{eq:combine2}
    \langle T''', D \rangle \leq \langle T', D \rangle + t \left ( \sum_{i=1}^n a_i  \right)
\end{equation}

\begin{equation}
\begin{split}
    \langle T''', D \rangle = \sum_{j=1}^n \sum_{i=1}^n T'''_{i, j} D_{i, j}
    &= \sum_{j=t+1}^n \sum_{i=1}^n T''_{i, j-t} |i-j| \\
    &= \sum_{j=t+1}^n \sum_{i=1}^n T''_{i, j-t} |i - (j-t) + t| \\
    & \leq \sum_{j=t+1}^n \sum_{i=1}^n T''_{i, j-t} |i - (j-t)|
    + t \sum_{j=t+1}^n \sum_{i=1}^n T''_{i, j-t} \\
    &= \sum_{j=1}^{n-t} \sum_{i=1}^n T''_{i, j} |i - j|
    + t \sum_{j=t+1}^n \sum_{i=1}^n T''_{i, j-t} \\
    &= \sum_{j=1}^{n} \sum_{i=1}^n T'_{i, j} |i - j|
    + t \sum_{j=t+1}^n \sum_{i=1}^n T''_{i, j-t} \\
    & \leq \left \langle T', D \right \rangle + t \mathbbm{1}^\top T' \mathbbm{1} \\
     & \leq \left \langle T', D \right \rangle + t \mathbbm{1}^\top (T' \mathbbm{1} + \tilde{a}) \qquad (\text{because } \tilde{a} \text{ is positive}) \\
     & = \left \langle T', D \right \rangle + t \mathbbm{1}^\top a
     \qquad (\text{row-sums constraint})
     \\
     &=
     \left \langle T', D \right \rangle + t \left ( \sum_{i=1}^n a_i \right )
\end{split}
\end{equation}
which shows that \cref{eq:combine2} holds. Now notice
\begin{equation}
\label{eq:combine3}
\begin{split}
    \widehat{W}_m(a, b) &= \langle T^\star, D(m) \rangle
    = \left \langle \left [ \begin{array}{cc}
    T' & \tilde{a} \\
    \tilde{b}^\top & c
    \end{array} \right ], \left [ \begin{array}{cc}
    D & m \mathbbm{1} \\
    m \mathbbm{1}^\top & 0
    \end{array} \right ] \right \rangle \\
    &= \langle T', D \rangle + m \sum_{i=1}^n \tilde{a}_i + m \sum_{i=1}^n \tilde{b}_i
    \end{split}
\end{equation}
so that combining \cref{eq:combine1,eq:combine2,eq:combine3} we have:
\begin{equation}
\begin{split}
   \widehat{W}(a, b') & \leq  \langle T''', D \rangle + m \sum_{i=1}^n \tilde{a}_i + m \sum_{i=1}^n \tilde{b}_i \\
   & \leq \langle T', D \rangle + t \left ( \sum_{i=1}^n a_i  \right)
   + m \sum_{i=1}^n \tilde{a}_i + m \sum_{i=1}^n \tilde{b}_i \\
   &= \widehat{W}_m(a, b)+ t \left ( \sum_{i=1}^n a_i  \right)
   \end{split}
\end{equation}
we have proven that \cref{eq:basic-ineq} holds, and the result follows as explained in \cref{eq:basic-ineq2,eq:basic-ineq3}

\subsection{Proof of \cref{lem:interp}}
We assume $m=1$. When $s=1$ we have
\begin{equation}
  A_1(i)\coloneqq \sum_{j=1}^i a_j - \sum_{j=1}^{i-1} a_j = a_i, \qquad B_s(i) \coloneqq \sum_{j=1}^i b_j - \sum_{j=1}^{i-1} b_j = b_i
\end{equation}
hence
\begin{equation}
\begin{split}
    \text{OTW}_{1, 1}(a, b) &\coloneqq | A_1(n) - B_1(n) |
	+ \sum_{i=1}^{n-1} |A_1(i) - B_1(i)| \\ &= | a_n - b_n | 
	+ \sum_{i=1}^{n-1} |a_i - b_i| = \|a-b\|_1 
	\end{split}
\end{equation}

Now when $s=n$ we have
\begin{equation}
  A_n(i) \coloneqq \sum_{j=1}^i a_j - \sum_{j=1}^{i-n} a_j = \sum_{j=1}^i a_j = A(i), \qquad B_n(i) \coloneqq \sum_{j=1}^i b_j - \sum_{j=1}^{i-n} b_j = \sum_{j=1}^i b_j = B(i)
\end{equation}
where this is due by convention: $i \leq n$ so that $i-n \leq 0$, and the sums from $j=1$ to $j=i-n \leq 0$ is an empty sum (so it has zero value). Then, we get
\begin{equation}
\begin{split}
    \text{OTW}_{1, n}(a, b) &\coloneqq | A_n(n) - B_n(n) |
	+ \sum_{i=1}^{n-1} |A_n(i) - B_n(i)| \\ &= | A(n) - B(n) | 
	+ \sum_{i=1}^{n-1} |A(i) - B(i)| = \text{OTW}_{m}(a,b) 
	\end{split}
\end{equation}

\end{document}